\documentstyle[aclap,epsf]{article}

\let\3=\ss\def\vm{{\sc Verbmobil}}
\def\dl#1#2{{\tt{}#1}\\({\it{}#2})\smallbreak}
\def\sa#1{\newline({\sc #1})}

\title{Insights into the Dialogue Processing of {\sc
Verbmobil}\thanks{Appears in the proceedings of ANLP-97, 31 March - 3
April, 1997, Washington, D.C.}}  \author{ Jan Alexandersson ~~~~\hfill
Norbert Reithinger ~~~~\hfill Elisabeth Maier \\ DFKI GmbH \\
Stuhlsatzenhausweg 3 \\ D-66123 Saarbr\"ucken, Germany \\
{\tt\{alexandersson,reithinger,maier\}@dfki.un-sb.de}}

\begin{document}

\maketitle
\vspace{-0.5in}
\begin{abstract} 

We present the dialogue module of the speech-to-speech translation
system \vm{}.  We follow the approach that the solution to dialogue
processing in a mediating scenario can not depend on a single
constrained processing tool, but on a combination of several simple,
efficient, and robust components. We show how our solution to dialogue
processing works when applied to real data, and give some examples
where our module contributes to the correct translation from German to
English.

\end{abstract}

\section{Introduction}\label{s:intro}

The implemented research prototype of the speech-to-speech translation
system \vm{} \cite{Wahlster93a,BubSchwinn96} consists of more than 40 modules for both speech and
linguistic processing.  The central storage for dialogue information
within the overall system is the dialogue module that exchanges data
with 15  of the other modules.

Basic notions within  \vm{} are {\em turns} and {\em utterances}. A turn
is defined as one contribution of a dialogue participant. Each turn
divides into utterances that sometimes resemble clauses as defined in a
traditional grammar. However, since we deal exclusively with spoken,
unconstrained contributions, utterances are sometimes just pieces of
linguistic material.

For the dialogue module, the most important dialogue related
information extracted for each utterance is the so called dialogue act
\cite{Jekatetal95}. Some dialogue acts describe solely the
illocutionary force, while other more domain specific ones describe
additionally aspects of the propositional content of an utterance. 

Prior to the  selection of the dialogue acts, we analyzed 
dialogues from \vm{}'s corpus of spoken and transliterated scheduling
dialogues. More than 500 of them have been annotated with dialogue
related information and serve as the empirical foundation of our work.

Throughout this paper we will refer to the example dialogue partly shown in
figure \ref{f:ex-dia}. The translations are as the deep processing line
of \vm{} provides them. We also annotated the utterances with the
dialogue acts as determined by the semantic evaluation module. {\tt
``//''} shows where utterance boundaries were determined.

We start with a brief introduction to dialogue processing in the \vm{}
setting.  Section \ref{s:context} introduces the basic data structures
followed by two sections describing some of the tasks which are
carried out within the dialogue module. Before the concluding remarks
in section \ref{s:conclusion}, we discuss aspects of robustness and
compare our approach to other systems.

\section{Introduction to  Dialogue Processing in \vm}\label{s:intro2}

\begin{figure*}[htb]
  {\parindent=0cm \small
\begin{center}
\begin{minipage}[t]{7.5cm}
\begin{flushleft}

\dl{{A01:} Tag // Herr Scheytt.
\sa{greet, introduce\_name}}{Hello, Mr Scheytt}

\dl{{B02:}
Guten Tag //
Frau Klein //
Wir m\"ussen noch einen Termin ausmachen // 
f\"ur die Mitarbeiterbesprechung.
\sa{greet, 
    introduce\_name, 
    init\_date, 
    suggest\_support\_date}
   }{Hello, Mrs. Klein, we should arrange an appointment, for the team
     meeting}

\dl{{A03:}
Ja,//
ich w\"urde Ihnen vorschlagen im Januar,// 
zwischen dem f\"unfzehnten und neunzehnten.
\sa{uptake,
    suggest\_support\_date,
    request\_comment\_date}
   }{Well, I would suggest in January, between the fifteenth and the
     nineteenth}
    
\dl{{B04:}
Oh // 
das ist ganz schlecht. //
zwischen dem elften und achtzehnten  Januar
bin ich in Hamburg.
\sa{uptake,
    reject\_date,
    suggest\_support\_date}
}{Oh,
that is really inconvenient,
I'm in Hamburg between the eighteenth of January and the eleventh,
}

{\large\bf \ldots}
\vspace{6pt}

\dl{{A09:}
Doch ich habe Zeit von sechsten Februar bis neunten Februar 
\sa{suggest\_support\_date}
}{I have time afterall from the 6th of February to the 9th of February}
  
\end{flushleft}
\end{minipage}~~~~\begin{minipage}[t]{7.5cm}
\begin{flushleft}
\dl{{B10:}
Sehr gut //
das pa\ss{}t bei mir auch //
Dann machen wir's gleich aus //
f\"ur Donnerstag //
den achten //
Wie w\"are es denn um acht Uhr drei\ss{}ig //
\sa{feedback\_acknowledgement,
    accept\_date,
    init\_date,
    suggest\_support\_date,
    suggest\_support\_date,
    suggest\_support\_date}
}{Very good,
that too suits me,
we will arrange for it,
for thursday,
the eighth,
how about half past eighth}

\dl{{A11:}
Am achten // ginge es bei mir leider nur bis zehn Uhr //
Bei mir geht es besser nachmittags .
\sa{suggest\_support\_date,
suggest\_support\_date,
accept\_date}
}{on the eighth,
Is it only unfortunately possible for me until 10 o'clock,
It suits me better in the afternoon
}

\dl{{B12:}
gut // um wieviel Uhr sollen wir uns dann treffen ?
\sa{feedback\_acknowledgement, suggest\_support\_date}
}{good,
when should we meet}

\dl{{A13:}
ich w"urde "ahm vierzehn Uhr vorschlagen // geht es bei Ihnen.
\sa{suggest\_support\_date,request\_comment\_date}}{
I would suggest 2 o'clock,
is that possible for you?}

\dl{{B14:}
sehr gut //
das pa"st bei mir auch //
das k"onnen wir festhalten 
\sa{accept\_date,accept\_date,accept\_date}
}{very good,
that suits me too,
we can make a note of that}

{\large\bf \ldots}
     
\end{flushleft}
\end{minipage}
  
\end{center}

} 
\caption{An example dialogue}\label{f:ex-dia}
\end{figure*}

In contrast to many other NL-systems, the \vm\ system is mediating a
dialogue between two persons. No restrictions are put on the
locutors, except for the limitation to stick to the approx.\ 2500 words
\vm{} recognizes. Therefore, \vm\ and
especially its dialogue component has to follow the dialogue in any
direction. In addition, the dialogue module is faced with incomplete and
incorrect input, and sometimes even gaps. 

When designing a component for such a scenario, we have chosen not to
use one big constrained processing tool. Instead, we have selected a
combination of several simple and efficient approaches, which together
form a robust and efficient processing platform.

As an effect of the mediating scenario, our module cannot serve as a
``dialogue controller'' like in man-machine dialogues. {The only
exception is  when  clarification dialogues are necessary between
\vm{} and a user.}

Due to its role as information server in the overall \vm{} system,
we started early in the project to collect requirements from other
components in the system. The result can be divided into three subtasks:

\begin{itemize}

        \item we allow for other components to {\em store} and {\em
        retrieve} context information. 

        \item we draw {\em inferences} on the basis of our input.

        \item we {\em predict} what is going to happen next.

\end{itemize}

Moreover, within \vm{} there are different processing tracks: parallel
to the deep, linguistic based processing, different shallow processing
modules also enter information into, and retrieve it from, the dialogue
module.  The data from these parallel tracks must be consistently
stored and made accessible in a uniform manner.

\begin{figure*}[htbp]
\begin{center}
\epsfxsize=135mm\leavevmode\epsfbox{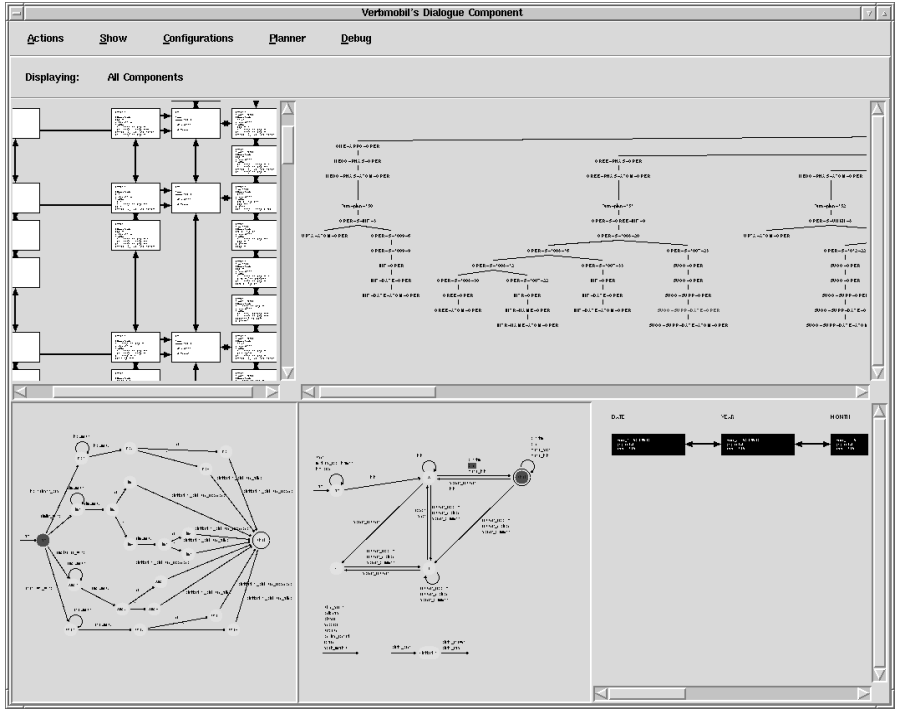}
\caption{Overview of the dialogue module}\label{f:all}
\end{center}
\end{figure*}

\begin{figure*}[bp]
\begin{center}
\epsfxsize=125mm\leavevmode\epsfbox{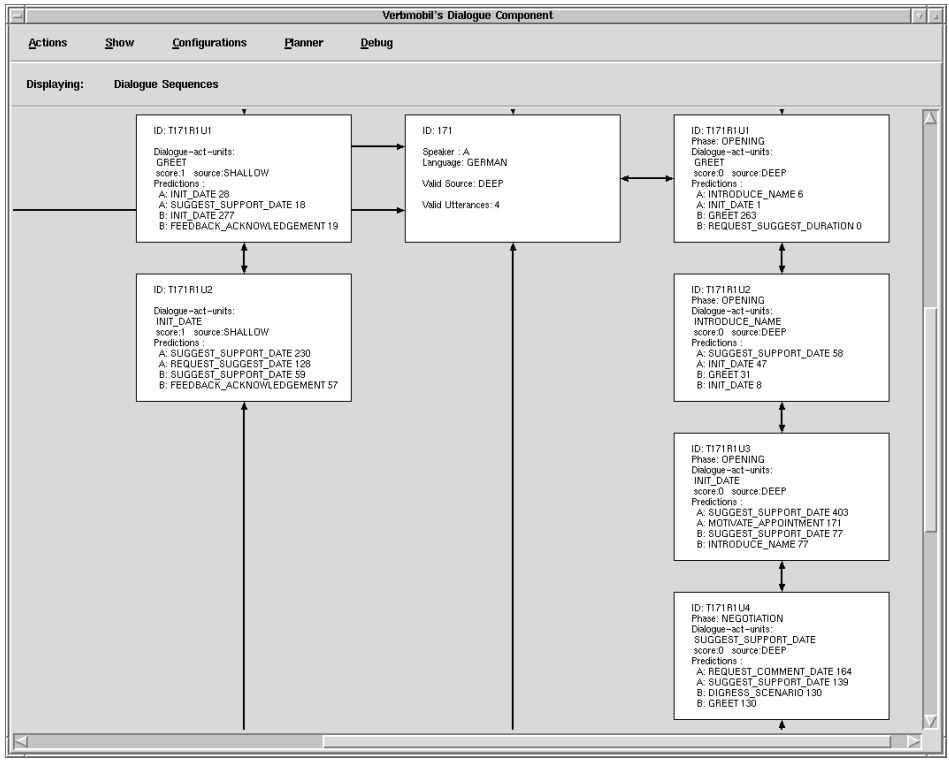}
\caption{A part of the sequence memory}\label{f:sequence}
\end{center}
\end{figure*}

Figure \ref{f:all} shows a screen dump of the graphical user interface
of our component while processing the example dialogue. In the upper
left corner we see the structures of the {\em dialogue sequence
memory}, where the middle right row represents {turns}, and the left
and right rows represent { utterances} as segmented by different
analysis components.  The upper right part shows the {\em intentional
structure} built by the plan recognizer. Our module contains two
instances of a {\em finite state automaton}. The one in the lower left
corner is used for performing clarification dialogues, and the other
for visualization purposes (see section \ref{s:related-work}). The
{\em thematic structure} representing temporal expressions is
displayed in the lower right corner.

\section{Maintaining  Context}\label{s:context}

As basis for storing context information we developed the {\em dialogue
sequence memory}. It is a generic structure which mirrors the
sequential order of turns and utterances.  A wide range of operation
has been defined on this structure.  For each turn, we store e.g.\ the
speaker identification, the language of the contribution, the
processing track finally selected for translation, and the number of
translated utterances. For the utterances we store e.g.\ the dialogue
act, dialogue phase, and predictions. These data are partly provided by
other modules of \vm{} or computed within the dialogue module itself
(see below).

Figure \ref{f:sequence} shows the dialogue sequence memory after the
processing of turn {\tt B02}. For the deep analysis side (to the
right), the turn is segmented into four utterances: {\em Guten Tag //
Frau Klein // Wir m\"ussen noch einen Termin ausmachen // f\"ur die
Mitarbeiterbesprechung}, for which the semantic evaluation component
has assigned  the dialogue acts {\sc Greet,
Introduce\_Name, Init\_Date,} and {\sc Suggest\_Support\_Date}
respectively.  To the left we see the
results of one of the shallow analysis components. It splits up the input into
two utterances {\em Guten Tag 
Frau Klein // Wir m\"ussen \ldots die
Mitarbeiterbesprechung} and assigns the dialogue acts
{\sc Greet} and {\sc Init\_Date}. 

The need for and use of this structure is highlighted by the following
example.  In the domain of appointment scheduling the German phrase
{\em Geht es bei Ihnen?} is ambiguous: {\em bei Ihnen} can either
refer to a location, in which case the translation is {\em Would it be
okay at your place?} or, to a certain time. In the latter case
the correct translation is {\em Is that possible for you?}. A simple
way of disambiguating this is to look at the preceding dialogue
act(s). In our example dialogue, turn {\tt A13}, the utterance {\em
ich w\"urde \"ahm vierzehn Uhr vorschlagen} ({\em I would hmm fourteen
o'clock suggest}) contains the proposal of a time, which is
characterized by the dialogue act {\sc suggest\_support\_date}. With
this dialogue act in the immediately preceding context the ambiguity is
resolved as referring to a time and the correct translation is
determined.

In our domain, in addition to the dialogue act the most important
propositional information are the dates as proposed, rejected, and
finally accepted by the users of \vm. While it is the task of the
semantic evaluation module to extract time information from the actual
utterances, the dialogue module integrates those information in its
{\em thematic memory}. This includes resolving relative time
expressions, e.g.\  {\em two weeks ago}, into precise time descriptions,
like ``23rd week of 1996''.
The information about the dates is split in a
specialization hierarchy. Each date to be negotiated serves as a root,
while the nodes represent the information about years, months,
weeks, days, days of week, period of day and finally time. Each node
contains also information about the attitude of the dialogue
participants concerning this certain item: proposed, rejected, or
accepted by one of the participants.

\begin{figure*}[htb]
\begin{center}
\epsfxsize=125mm\leavevmode\epsfbox{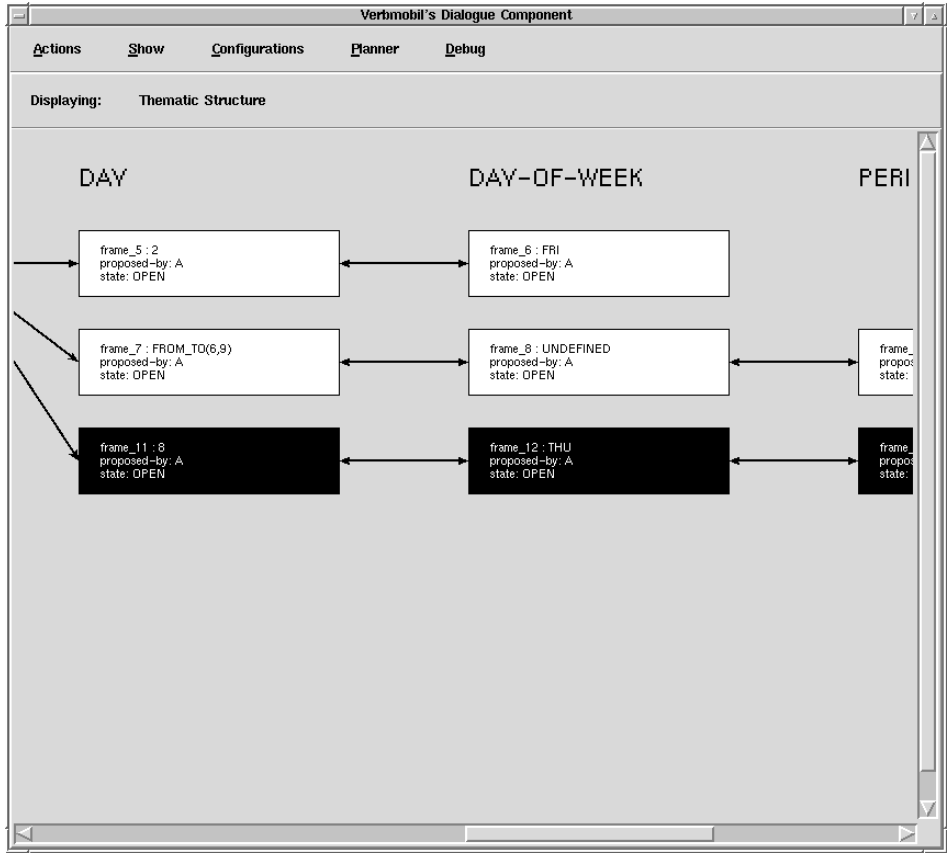}
\caption{Day/Day-of-Week detail of the thematic structure}\label{f:thema}
\end{center}
\end{figure*}

Figure \ref{f:thema} shows parts of the thematic structure after the
processing of turn {\tt B10}. The black boxes stand for the date
currently under consideration. Thursday, 8., is the current date
agreed upon. We also see the previously proposed interval from
6.-9. of the same month in the box above ({\tt FROM\_TO(6,9)}).

\begin{figure*}[htb]
\begin{center}
\epsfxsize=135mm\leavevmode\epsfbox{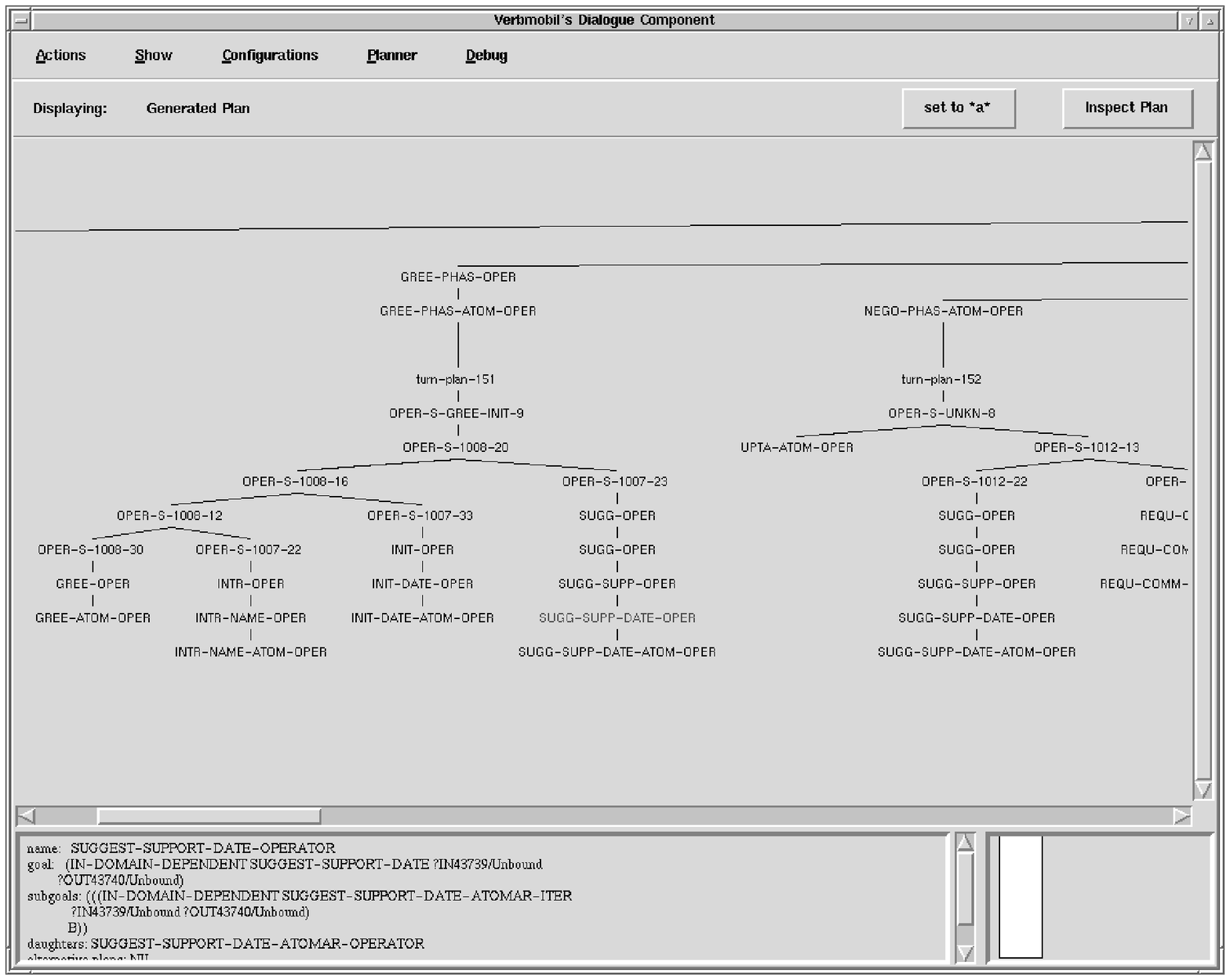}
\caption{Intentional structure for two turns}\label{f:plan}
\end{center}
\end{figure*}

\section{Inferences}\label{s:inferences}

Besides the mere storage of dialogue related data, there are also
inference mechanisms  integrating the data in representations of
different aspects of the dialogue. These data are again stored in the
context memories shown above and are accessed by the other \vm{}
modules.

\subsubsection*{Plan Based Inferences}

Inspecting our corpus, we can distinguish three phases in most of the
dialogues. In the first, {\em the opening phase}, the locutors greet
each other and the topic of the dialogue is introduced. The dialogue
then proceeds into {\em the negotiation phase}, where the actual
negotiation takes place. It concludes in {\em the closing phase} where
the negotiated topic is confirmed and the locutors say goodbye. This
phase information  contributes to the correct transfer of an
utterance. For example, the German utterance {\em Guten Tag} is
translated to ``Hello'' in the greeting phase, and to ``Good day'' in
the closing phase.

The task of determining the phase of the dialogue has been given to the
plan recognizer \cite{Alexandersson95b}. It builds a tree like
structure which we call the intentional structure.  The current version
makes use of plan operators both hand coded and automatically derived
from the \vm{} corpus. The method used is transferred from the field of
grammar extraction \cite{Stolcke94a}. To contribute to the
robustness of the system, the processing of the recognizer is divided
into several processing levels like the ``turn level'' and the ``domain
dependent level''. The concepts of turn levels and the automatic
acquisition of operators are described in \cite{Alexandersson96}.

In figure \ref{f:plan} we see the structure after processing turns {\tt
B02} and {\tt A03}.  The leaves of the tree are the dialogue acts. The
root node of the left subtree for {\tt B02} is a {\tt
GREE(T)-INIT-...} operator which belongs to the greeting phase,
while the partly visible one to the right belongs to the negotiation
phase.

In the example used in this paper we are processing a ``well formed''
dialogue, so the turn structure can be linked into a structure spanning
over the whole dialogue.  We also see in figure \ref{f:sequence} how
the phase information has been written into the boxes representing the
utterances  of turn {\tt B02} as segmented by the deep analysis.

\subsubsection*{Thematic Inferences}

In scheduling dialogues, referring expressions like the German word
{\em n\"achste} occur frequently.  
Depending on the thematic structure it can be translated as {\em next}
if the date referred to is immediately after the speaking time, or
{\em following} in the other cases.
The thematic structure is mainly used
to resolve this type of anaphoric expressions if requested by the
semantic evaluation or the transfer module. The information about the
relation between the date under consideration and the speaking time
can be immediately computed from the thematic structure.

The thematic structure is also used to check whether the time
expressions are correctly recognized. If some implausible dates are
recognized, e.g. April, 31., a clarification can be invoked. The
system proposes the speaker a more plausible date, and waits for an acceptance or
rejection of the proposal. In the first case, the correct date will be
translated, in the latter, the user is asked to repeat the whole
turn.

Using the current state of the thematic structure and the dialogue act
in combination with the time information of an utterance, multiple
readings can be inferred \cite{Maier96a}. For example, if both locutors
propose different dates, an implicit rejection of the former date can
be assumed.

\section{Predictions}\label{s:predictions}

A different type of inference is used to generate predictions about
what comes next. While the plan-based component uses declarative
knowledge, albeit acquired automatically, dialogue act predictions are
based solely on the annotated \vm{} corpus. The computation uses the
conditional frequencies of dialogue act sequences to compute
probabilities of the most likely follow-up dialogue acts
\cite{Reithingeretal96}, a method adapted from language modeling \cite{Jelinek90}. As described above, the dialogue sequence memory
serves as the central repository for this information.

The sequence memory in figure \ref{f:sequence} shows in addition to
the actual recognized dialogue act also the predictions for the
following utterance. In \cite{Reithingeretal96} it is demonstrated
that exploiting the speaker direction significantly enhances the
prediction reliability. Therefore, predictions are computed for both
speakers. The numbers after the predicted dialogue acts show the
prediction probabilities times 1000.

As can be seen in the figure, the actually recognized dialogue acts
are, for this turn, among the two most probable predicted
acts. Overall, approx.\ 74\% of all recognized dialogue acts are
within the first three predicted ones.

Major consumers of the predictions are  the semantic evaluation
module, and the shallow translation module. The former module that
uses mainly knowledge based methods to determine the dialogue act of
an utterance exploits the predictions to narrow down the number of
possible acts to consider. The shallow translation module integrates the
predictions within a Bayesian classifier to compute dialogue acts
directly from the word string.

\section{Robustness}\label{s:robustness}

For the dialogue module there are two major points of insecurity
during  operation. On the one hand, the user's dialogue behaviour
cannot be controlled. On the other hand, the segmentation as computed
by the syntactic-semantic construction module, and the dialogue acts as
computed by the semantic evaluation module, are very often not the
ones a linguistic analysis on the paper will produce. Our example
dialogue is a very good example for the latter problem.

Since no module in \vm{} must ever crash, we had to apply various
methods to get a high degree of robustness.  The most knowledge
intensive module is the plan recognizer. The robustness of this
subcomponent is ensured by dividing the construction of the intentional
structure into several processing levels. Additionally, at the turn
level the operators are learned from the annotated corpus.  If the
construction of parts of the structure fails, some
functionality has been developed to recover. An important ingredience
of the processing is the notion of {\em repair} -- if the plan
construction is faced with something unexpected, it uses a set of
specialized repair operators to recover.  If parts of the structure
could not be built, we can estimate on the basis of predictions what the
gap consisted of.

The statistical knowledge base for the prediction algorithm is trained
on the \vm{} corpus that in its major parts contains well-behaved
dialogues. Although prediction quality gets worse if a sequence of
dialogue acts has never been seen, the interpolation approach to
compute the predictions still delivers useful data.

As mentioned above, to contribute to the correctness of the overall
system we perform different kinds of clarification dialogues with the
user. In addition to the inconsistent dates, we also e.g.\ recognize
similar words in the input that will be most likely exchanged by the
speech recognizer. Examples are the German words for {\em
thirteenth\/} ({\em dreizehnter\/}) and {\em thirtieth\/} ({\em
drei\ss{}igster\/}). Within a uniform computer--human interaction, we
resolve these problems.

\section{Related Work}\label{s:related-work}

In the speech-to-speech translation system {\sc Janus}
\cite{Lavieetal96a}, two different approaches, a plan based and an
automaton based, to model dialogues have been implemented. Currently,
only one is used at a time. For \vm{},
\cite{AlexanderssonReithinger95} showed that the descriptive power of
the plan recognizer and the predictive power of the statistical
component makes the automaton obsolete. 

The automatic acquisition of a dialogue model from a corpus is reported
in \cite{Kitaetal96}. They extract a probabilistic automaton using an
annotated corpus of up to 60 dialogues. The transitions correspond to
dialogue acts. This method captures only local discourse structures,
whereas the plan based approach of \vm{} also allows for the description
of global structures.
Comparable structures are also defined in the dialogue processing of
{\sc Trains} \cite{TraumAllen92}. However, they are defined manually
and have not been tested on larger data sets.

\section{Conclusion and Future Work}\label{s:conclusion}

Dialogue processing in a speech-to-speech translation system like \vm{}
requires innovative and robust methods. In this paper we presented
different aspects of the dialogue module while processing one example
dialog.  The combination of knowledge based and statistical methods
resulted in a reliable system. Using the \vm{} corpus as empirical
basis for training and test purposes significantly improved the
functionality and robustness of our module, and allowed for focusing our
efforts on real problems. The system is fully integrated in the \vm{}
system and has been tested on several thousands of utterances.

Nevertheless, processing in the real system creates still new
challenges. One problem that has to be tackled in the future is the
segmentation of turns into utterances. Currently, turns are very often
split up into too many and too small utterances. In the future, we
will have to focus on the problem of ``glueing'' fragments
together. When given back to the transfer and generation modules, this
will enhance translation quality.

Future work includes also more training and the ability to handle
sparse data. Although we use one of the largest annotated corpora
available, for purposes like training we still need more data.

\section*{Acknowledgements}

{This work was funded by the German Federal Ministry of Education,
        Science, Research and Technology (BMBF) in the framework of the
        \vm{} Project under Grant {01IV101K/1}. The responsibility for
        the contents of this study lies with the authors.  We thank our
        students Ralf Engel, Michael Kipp, Martin Klesen, and Paula
        Sevastre for their valuable contributions. Special thanks to
        Reinhard for Karger's Machine.}

\end{document}